\documentclass[runningheads]{llncs}
\usepackage[T1]{fontenc}
\usepackage{graphicx}
\usepackage{amsmath}
\usepackage{cleveref}
\begin{document}
\title{A Synthetic Benchmarking Pipeline to Compare Camera Calibration Algorithms}
\titlerunning{SynthCal}
%
\author{Lala Shakti Swarup Ray\inst{1}\orcidID{0000-0002-7133-0205} \and
Bo Zhou\inst{1,2}\orcidID{0000-0002-8976-5960} \and
Lars Krupp\inst{1,2}\orcidID{0000-0001-6294-2915} \and
Sungho Suh\inst{1,2}\orcidID{0000-0003-3723-1980}\and
Paul Lukowicz\inst{1,2}\orcidID{0000-0003-0320-6656} }
\authorrunning{Ray et al.}
%
\institute{German Research Center for Artificial Intelligence , Kaiserslautern, Germany \\
\and
RPTU Kaiserslautern-Landau, Kaiserslautern, Germany\\}
\maketitle              
\begin{abstract}
Accurate camera calibration is crucial for various computer vision applications. However, measuring calibration accuracy in the real world is challenging due to the lack of datasets with ground truth to evaluate them. In this paper, we present SynthCal, a synthetic camera calibration benchmarking pipeline that generates images of calibration patterns to measure and enable accurate quantification of calibration algorithm performance in camera parameter estimation. We present a SynthCal-generated calibration dataset with four common patterns, two camera types, and two environments with varying view, distortion, lighting, and noise levels for both monocular and multi-camera systems. The dataset evaluates both single and multi-view calibration algorithms by measuring re-projection and root-mean-square errors for identical patterns and camera settings. Additionally, we analyze the significance of different patterns using different calibration configurations. The experimental results demonstrate the effectiveness of SynthCal in evaluating various calibration algorithms and patterns. 

\end{abstract}
\section{Introduction}
\label{sec:intro}
When we capture an image using a camera, the captured digital image can differ from the real-world scene in terms of perspective, distortion, color, resolution, and other visual properties. This is because real-world scenes are three-dimensional and continuous, while digital images captured by a camera are two-dimensional and discrete, and contain distortion and other imperfections. To minimize these differences and improve the accuracy of image-based computer vision tasks, camera calibration is essential. 

Camera calibration involves calculating camera parameters that refer to its intrinsic and extrinsic characteristics for accurately mapping points in the 3D world to their corresponding 2D image coordinates. 
Once the camera is calibrated, it can accurately measure distances, angles, and sizes of objects in the 3D world and perform other image-based computer vision tasks such as object tracking \cite{zhang2023motrv2}, 3D reconstruction \cite{kang2020review,wu2023multiview}, medical imaging \cite{barbero2019smartphone}, and autonomous driving \cite{feng2019can}. 

Geometric camera calibration \cite{kikkawa2023accuracy,huai2023review} is one of the most widely used calibration methods. It involves using a calibration target with known geometric features, such as a calibration grid, to estimate the camera parameters. 

However, creating real camera calibration data with ground truth for calibration algorithms can be challenging because it is difficult to measure camera position and rotation accurately, and the camera's intrinsic parameters can change with changes in the zoom level, focus distance, or temperature.   
Moreover, cameras can have different intrinsic parameters, even if they are of the same make and model, because of manufacturing tolerances, assembly errors, and differences in lens quality.
Observing the calibration pattern in the image along with the previous knowledge of the pattern, we can determine the intrinsic and extrinsic parameters using various calibration algorithms, such as Zhang \cite{zhang2000flexible}, Tsai \cite{tsai1987versatile}, or Bouguet calibration method \cite{bouguet2004camera}.
Previous works have tried to compare different camera calibration algorithms \cite{usamentiaga2018comparison}. However, there is a need for a benchmarking procedure that can provide a quantitative comparison of calibration algorithms due to the unknown ground truth of the calibration dataset.

\begin{figure*}[!t]
\begin{center}
\includegraphics[width=\linewidth]{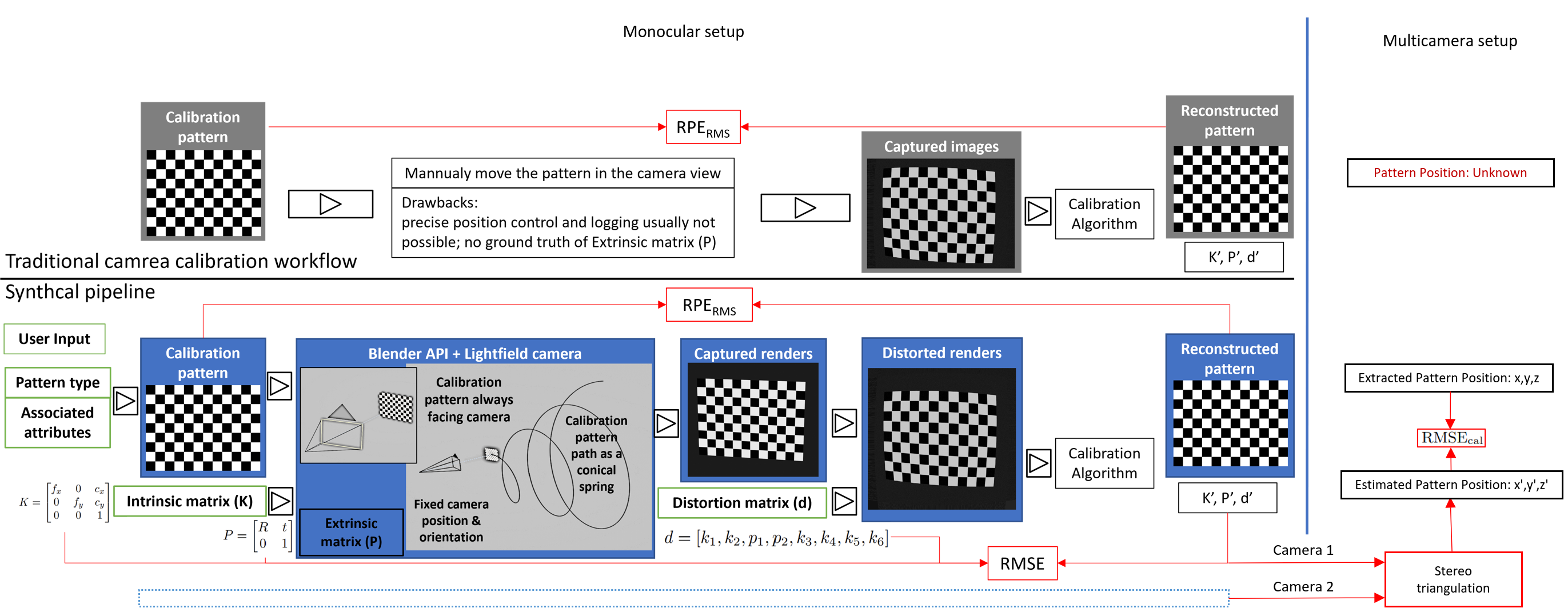}
\end{center}
   \caption{SynthCal pipeline to generate calibration dataset from a set of input attributes: Pattern attributes, camera intrinsic, distortion, extrinsic matrix. The accuracy is then evaluated using $RMSE$ and $RPE_{RMS}$ for monocular cameras.
   }
\label{fig:1}
\end{figure*}

To overcome these problems, we introduce an overall pipeline, named SynthCal, which generates a synthetic camera calibration dataset with user-defined intrinsic camera parameters while precisely measuring the extrinsic camera parameters. It enables the selection of the optimal camera calibration algorithm for specific configurations by considering all intrinsic, extrinsic, and distortion parameters.
Additionally, it ensures that lighting conditions and noise are identical for the different captured datasets for accurate comparison of different calibration patterns which is not possible in the real world.
The idea of generating synthetic calibration data has been previously applied in other works, such as sports-based synthetic calibration \cite{chen2019sports} and evaluating closed-form solutions of principal line calibration \cite{chuang2021geometry}, but not necessarily for comparing calibration algorithms.

Our main contributions can be summarized as follows:
\begin{itemize}
    \item We present a pipeline to generate a camera calibration dataset with ground truth parameters and select the optimal camera calibration algorithm for the specific configurations, as depicted in \cref{fig:1}. 
    \item We validate the proposed pipeline on three different camera calibration algorithms which is consistent with previous works and then use SynthCal-generated dataset to compare four different calibration patterns given in \cref{fig:3} for monocular and multi-camera systems with two distinct camera configurations, and two different lighting and noise conditions.
\end{itemize}
\section{Proposed Method}
\label{sec:proposedmthod}

\begin{figure}[!t]
\begin{center}
\includegraphics[width=0.9\linewidth]{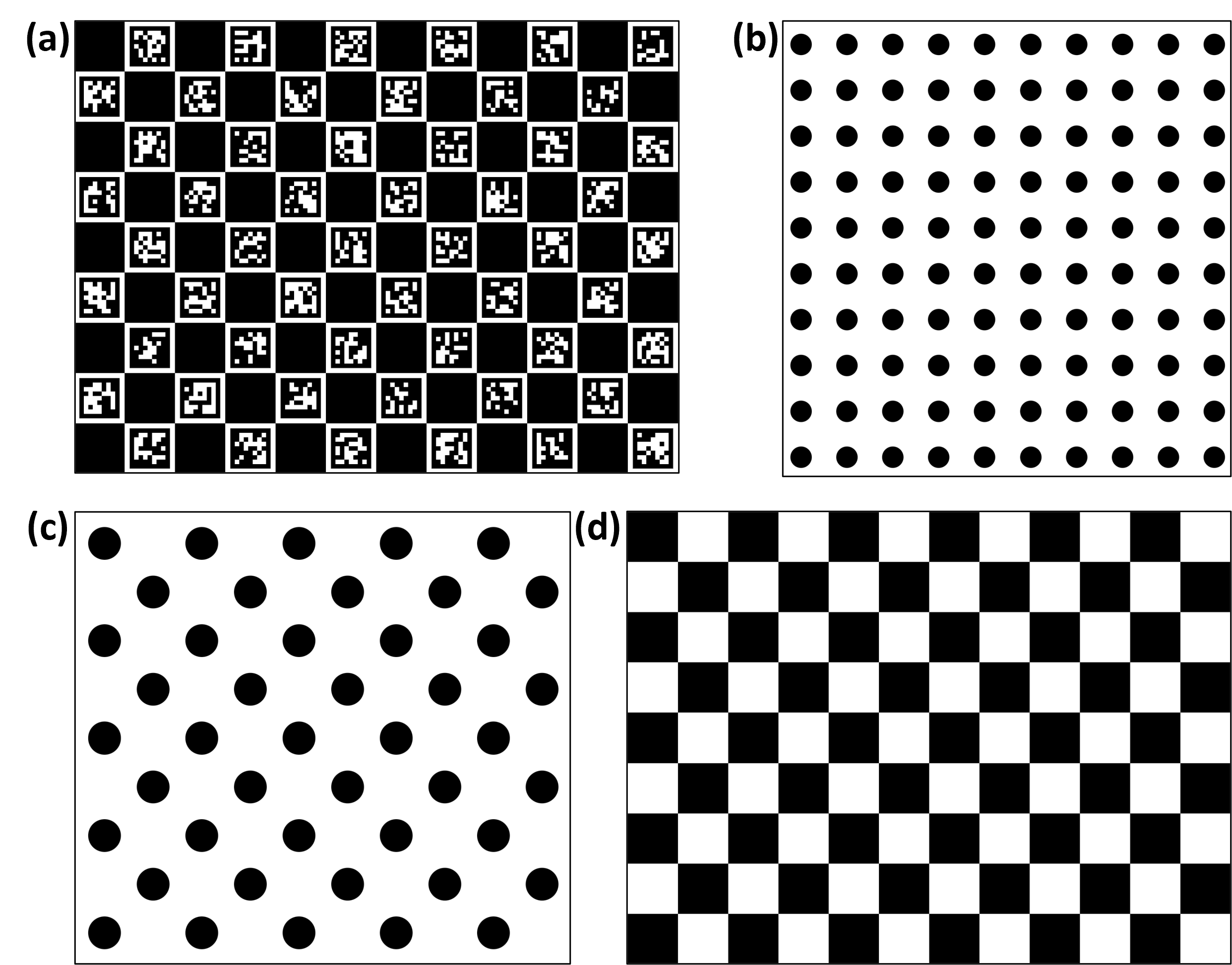}
\end{center}
   \caption{(a) 9 $\times$ 12 Charuco pattern, (b) 10 $\times$ 10 Symmetric circle grid, (c) 9 $\times$ 10 Asymmetric circle grid, (d) 9 $\times$ 12 Checkerboard pattern.}
\label{fig:3}
\end{figure}

We created a modular web-based interface with OpenCV and Blender API in the back-end to generate a synthetic camera calibration dataset with ground truth which has functionalities to create different camera calibration patterns, simulate a camera inside Blender using the light-field analysis add-on \cite{honauer2017dataset}, render the camera calibration pattern from various positions and orientations, add radial distortions while establishing the camera's intrinsic, extrinsic and distortion parameters to formulate the ground truth.
We used an OpenCV to generate geometric patterns that take input pattern type, and pattern attributes to generate a PNG image. Our script allows us to create checkerboard patterns (Ch), symmetric circular patterns (Sc), asymmetric circular patterns (Ac), and Charuco \cite{an2018charuco} patterns (Cu) of different configurations as shown in \cref{fig:3}.
Let $K$ be the intrinsic matrix of the camera, which includes the parameters that describe the internal configuration of the camera, such as the focal length ($f_x$, $f_y$) and principal point ($c_x$, $c_y$):
\begin{equation}
K = \begin{bmatrix} f_x & 0 & c_x \\ 0 & f_y & c_y \\ 0 & 0 & 1 \end{bmatrix}
\end{equation}
We used a Blender python API and a light-field add-on to create synthetic cameras that take camera attributes ($f_x$, $f_y$) and ($c_x$, $c_y$) to create a camera configuration file for simulating the camera inside Blender.
To capture the calibration pattern for dataset creation, we moved the pattern in a path resembling the shape of a conical spring, as depicted in \cref{fig:1}. The center of the calibration pattern is always in the camera's direction, so the planar pattern can be captured in different angles, sizes, and orientations and have consistency without going out of the camera frame. 
Let $R$ be the rotation matrix that describes the orientation of the camera in the global coordinate system, and let $t$ be the translation vector that describes the position of the camera in the world coordinate system:
\begin{equation}
P = \begin{bmatrix} R & t \\ 0 & 1 \end{bmatrix}
\end{equation}
The extrinsic matrix $P$ combines the rotation matrix and the translation vector.
$R$ is, and $t$ are evaluated by extracting the global position and orientation of the camera and calibration pattern at each frame.
The camera parameters can also be described using the distortion parameters, which describe the deviations from the ideal imaging system. The distortion parameters can be represented as a vector 
\begin{equation}
    d = [k_1, k_2, p_1, p_2, k_3, k_4, k_5, k_6]
\end{equation}

\begin{figure}[!t]
\begin{center}
\includegraphics[width=0.9\linewidth]{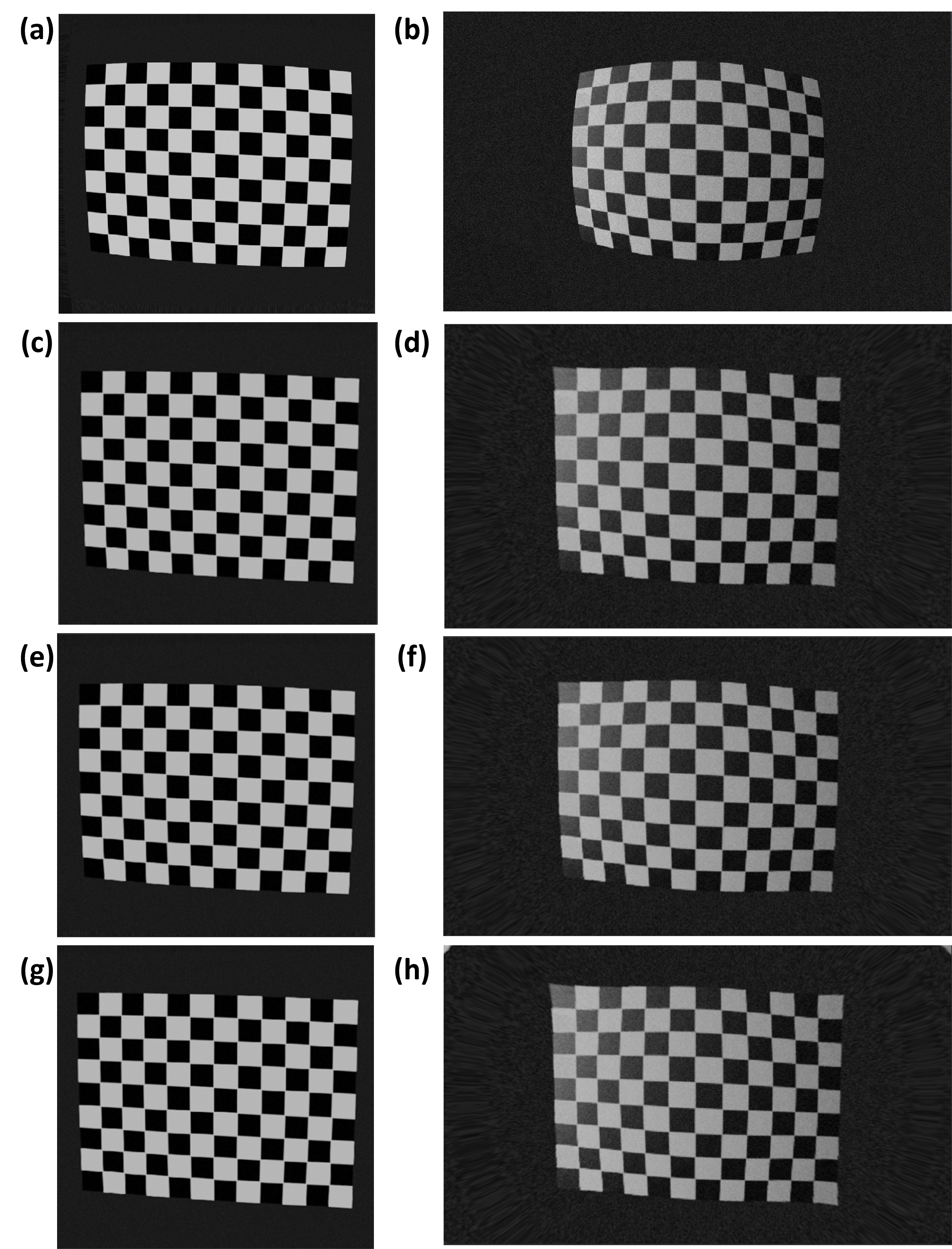}
\end{center}
   \caption{(a, b) Original clean and noisy capture, undistorted render using the camera parameters predicted by (c, d) Zhang, (e, f) Tsai, (g, h) Bouguet method.}
\label{fig:4}
\end{figure}

where $k_1, k_2, k_3, k_4, k_5, k_6$ are radial distortion coefficients and $p_1, p_2$ are tangential distortion coefficients.
The distortions are added later using Blender undistorted node by setting up a tracking scene in Blender and defining $K$ and $d$.
The final equation for mapping $X$ a 3D point in the global coordinate system to $x$ a 2D point in the image plane, including the distortion parameters, can be written as:
\begin{equation}
x = K [R \mid t] X + d(\frac{x_d}{f_x}, \frac{y_d}{f_y})    
\end{equation}
Where $x_d$ and $y_d$ are the distorted image coordinates, the distortion model $d$ maps the distorted image coordinates to the corrected image coordinates.
The captures are saved in PNG formats, while camera parameters are saved as NumPy arrays.

For a multicamera system, the synthetic dataset generation process involves considering the intrinsic matrices of multiple cameras and simulating their interactions. In this scenario, let's denote the intrinsic matrices of the two cameras as \(K_1\) and \(K_2\) respectively, with corresponding distortion parameters \(d_1\) and \(d_2\). The extrinsic matrices \(P_1\) and \(P_2\) represent the position and orientation of the cameras in the global coordinate system.

To extend the methodology for a multicamera setup, the calibration pattern is moved along a trajectory that ensures visibility from both cameras. The conical spring-like path is designed to capture the pattern from varying angles, sizes, and orientations for each camera while maintaining consistency. The rotation matrices \(R_1\) and \(R_2\), as well as translation vectors \(t_1\) and \(t_2\), are determined individually for each camera frame.

The distortion parameters \(d_1\) and \(d_2\) are applied separately to the distorted image coordinates \(x_{d1}\) and \(x_{d2}\) of each camera. The final mapping equation for a point \(X\) in the global coordinate system to its respective distorted image coordinates \(x_1\) and \(x_2\) in the image planes of Camera 1 and Camera 2 is given by:

\begin{equation}
\begin{bmatrix} x_{1} \\ x_{2} \end{bmatrix} = \begin{bmatrix} K_{1} [R_{1} \mid t_{1}] \\ K_{2} [R_{2} \mid t_{2}] \end{bmatrix} X + \begin{bmatrix} d_{1}\left(\frac{x_{d1}}{f_{x1}}, \frac{y_{d1}}{f_{y1}}\right) \\ d_{2}\left(\frac{x_{d2}}{f_{x2}}, \frac{y_{d2}}{f_{y2}}\right) \end{bmatrix}
\end{equation}

In this formulation, the intrinsic matrices \(K_1\) and \(K_2\) encapsulate the parameters specific to each camera, while the distortion parameters \(d_1\) and \(d_2\) account for individual radial and tangential distortions. The resulting synthetic dataset includes images from both cameras, with their respective intrinsic and extrinsic parameters saved for each frame in the dataset.

\section{Results}
\label{sec:experiments}
\subsection{Dataset}
We created a dataset of four widely used distinct pattern types that are a 9$\times$12 checkerboard pattern with a checker width of 15 mm, one 10$\times$10 symmetric circle pattern with a 7 mm circle diameter, and 15 mm circle spacing, one 9$\times$10 asymmetric circle pattern with 9 mm diameter, and 22 mm diagonal spacing and 9$\times$12 Charuco pattern checker width of 15 mm and ArUco dictionary \cite{tzortzis2014minmax} of 7$\times$7.
Two distinct camera configurations representing a high-resolution rectilinear lens with focal length ($3000$, $3000$), principal point ($2048$, $1536$) with distortion parameters $[ 0.05, 0.02, 0.001, 0, 0, 0, 0, 0] $ and a low resolution wide, angle lens with focal length ($600$, $450$), principal point ($320$, $240$) with distortion parameters $[ 0.5, 0.1, 0.03, 0, 0, 0, 0, 0]$ are simulated for capturing the patterns. Multiple cameras with either camera configuration are added to the scene to create a stereo dataset.
Skew and tangential distortion are kept at zero for all camera configurations.
The extrinsic parameters $R$ a 3$\times$3 identity matrix and $t$  a 3$\times$1 zero vector are calculated using vector calculation with the relative position and orientation of the camera and target pattern to establish the ground truth. 
Two different external lighting conditions are used while rendering, one with uniform light across the scene without noise (Clean) and another with Directional lights with additive Gaussian noise (Noisy) in the camera captures.
We created 40 data configurations with 127 captures with camera intrinsic and extrinsic matrix for each configuration with mono and stereo settings as specified in \cref{tab:table1}.
\begin{table}[!t]
  \begin{center}
  \caption{Dataset statistics specifying eight different configurations based on two camera types, four pattern types, and two different environment factors available in the dataset.}
  \label{tab:table1}
    {\small{
\begin{tabular}{c|c|c}
\hline
Camera & Pattern & Environment \\
\hline
Rectilinear lens & 9$\times$12 Ch, 10$\times$10 Sc, & Clean\\
(mono + stereo)& 9$\times$10 Ac, 9$\times$12 & Noisy\\
\hline
Wide angle lens & 9$\times$12 Ch, 10$\times$10 Sc, & Clean\\
(mono + stereo)& 9$\times$10 Ac, 9$\times$12 Cu & Noisy\\
\hline
Rectilinear + Wide angle & 9$\times$12 Ch, 10$\times$10 Sc, & Clean\\
(stereo)& 9$\times$10 Ac, 9$\times$12 Cu & Noisy\\
\hline
\end{tabular}
}}
\end{center}
\end{table}

\subsection{Evaluation}
\begin{table}[!t]
  \begin{center}
  \caption{ $\mathrm{RPE_{RMS}}$ and $\mathrm{RMSE}$ calculated for different lens using different camera calibration methods for 9$\times$12 Ch.}
  \label{table2}
    {\small{
\begin{tabular}{c|c|c|c}
\hline
Camera & Algorithm & \textbf{$\mathrm{RPE_{RMS}}$} & \textbf{$\mathrm{RMSE}$} \\
\hline
& Zhang's
method & 0.510 & 1.221\\
Rectilinear lens. & Tsai's method & 0.551 & 1.880\\
& Bouguet method & \textbf{0.373} & \textbf{1.127}\\
\hline
& Zhang's
method & 1.316 & 2.219\\
Wide angle lens. & Tsai's method & 1.433 & 2.344\\
& Bouguet method & \textbf{0.811} & \textbf{1.861}\\
\hline
\end{tabular}
}}
\end{center}
\end{table}

\begin{table*}[!t]
  \begin{center}
  \caption{ $\mathrm{RPE_{RMS}}$ and $\mathrm{RMSE}$ calculated for four different calibration patterns in two different environmental conditions.}
  \label{table3}
    {\small{
\begin{tabular}{c|c|c|c|c|c|c|cc}
\hline
Environment & Pattern & \multicolumn{2}{c}{Zhang's Method} & \multicolumn{2}{|c|}{Tsai's Method} & \multicolumn{2}{c}{Bouguet's Method} \\
\cline{3-8}
 &  & \textbf{$\mathrm{RPE_{RMS}}$} & \textbf{$\mathrm{RMSE}$} & \textbf{$\mathrm{RPE_{RMS}}$} & \textbf{$\mathrm{RMSE}$} & \textbf{$\mathrm{RPE_{RMS}}$} & \textbf{$\mathrm{RMSE}$} \\
\hline
Clean & 9$\times$12 Ch & 0.510 & 1.221 & 0.561 & 1.233 & \textbf{0.483} & \textbf{1.166} \\
 & 10$\times$10 Sc & 0.508 & 1.206 & 0.559 & 1.216 & \textbf{0.490} & \textbf{1.214} \\
& 9$\times$10 Ac & 0.506 & 1.205 & 0.554 & 1.198 & \textbf{0.497} & \textbf{1.193} \\
& 9$\times$12 Cu & 0.493 & 1.093 & 0.542 & 1.139 & \textbf{0.476} & \textbf{1.012} \\
\hline
Noisy & 9$\times$12 Ch & 1.116 & \textbf{2.219} & 1.227 & 2.290 & \textbf{1.062} & 2.252 \\
 & 10$\times$10 Sc & 1.20 & \textbf{2.263} & 1.320 & 2.347 & \textbf{1.140} & 2.373 \\
& 9$\times$10 Ac & 1.118 & \textbf{2.261} & 1.234 & 2.224 & \textbf{1.155} & 2.319 \\
& 9$\times$12 Cu & 0.898 & \textbf{1.916} & 0.987 & 1.868 & \textbf{0.853} & 2.020 \\
\hline
\end{tabular}
}}
\end{center}
\end{table*}

\begin{table*}[!t]
  \begin{center}
  \caption{ $\mathrm{RMSE_{cal}}$ calculated over the global position of the calibration pattern for different multi-camera systems for 9$\times$12 Ch using the triangulation method.}
  \label{table4}
    {\small{
\begin{tabular}{c|c|c|c|c}
\hline
&& Zhang's Method & Tsai's Method & Bouguet's Method \\
\hline
Cameras & Environment &  \multicolumn{3}{c}{\textbf{$\mathrm{RMSE_{cal}}$}} \\
\hline
2$\times$Rectilinear lens& Clean & 2.365 & 2.456 & \textbf{2.143}\\
& Noisy & \textbf{4.604} & 4.812 & 4.746\\
\hline
2$\times$Wide angle lens& Clean & 3.780 & 3.998 & \textbf{3.612}\\
& Noisy & \textbf{5.611} & 5.742 & 5.798\\
\hline
Rectilinear \& & Clean & 4.118 & 4.236 & \textbf{4.052}\\
wide angle lens & Noisy & \textbf{6.401} & 6.552 & 6.577\\
\hline
\end{tabular}
}}
\end{center}
\end{table*}

We used $\mathrm{RMS\ Reprojection\ Error}$ ($\mathrm{RPE_{RMS}}$) as a metric to compare the algorithms and calibration patterns which can be defined as:
\begin{equation}
    \mathrm{RPE_{RMS}} = \sqrt{\frac{1}{N}\sum_{i=1}^{N} \left\| \mathbf{x}_i - \mathbf{\hat{x}}_i \right\|^2}
\end{equation}
where $N$ is the number of points, $\mathbf{x}_i$ is the observed image point in the captured image, and $\mathbf{\hat{x}}_i$ is the corresponding projected image point using the estimated intrinsic and extrinsic parameters from the camera calibration.
We also calculated accuracy by comparing the estimated intrinsic and extrinsic parameters of the camera to the ground truth values using $\mathrm{Root\ Mean\ Square \ Error}$ ($\mathrm{RMSE}$) that can be defined as:
\begin{equation}
    \mathrm{\mathrm{RMSE}} = \sqrt{\frac{1}{L} \sum_{i=1}^{L} \left( X_i - \hat{X}_i \right)^2}
\end{equation}
where $L$ is the number of parameters being estimated, $X_i$ is the ground truth value for the $i$-th parameter, and $\hat{X}_i$ is the estimated value for the $i$-th parameter.

In our multicamera setup, we employed stereo triangulation \cite{hahne2018baseline} using the following equation to calculate the global position (\(X\)) of a calibration pattern, assuming knowledge of the camera parameters for two cameras within the system:

\begin{equation}
X = \frac{(X_1 - t_{1}) \times (X_2 - t_{2})}{\| (X_1 - t_{1}) \times (X_2 - t_{2}) \|}
\end{equation}

Here, \(X_1\) and \(X_2\) represent the 3D points in the coordinate systems of Camera 1 and Camera 2, respectively, and \(t_{1}\) and \(t_{2}\) are the translation vectors of Camera 1 and Camera 2. Subsequently, we utilized the Root Mean Square Error (RMSE) to quantify the disparity between the calculated global positions and the corresponding positions extracted from a simulation. The RMSE equation for a set of XYZ points involves calculating the square root of the average of the squared differences between the simulated XYZ coordinates (\(X_{sim}\)) and the calculated XYZ coordinates (\(X_{calc}\)):

\begin{equation}
\mathrm{RMSE_{cal}} = \sqrt{\frac{1}{N} \sum_{i=1}^{N} \left((X_{sim_i} - X_{calc_i})^2\right)}
\end{equation}

This facilitated a comprehensive evaluation of the accuracy of our global position calculations in comparison to simulated ground truth, providing a quantitative measure through the \(RMSE_{cal}\) metric (which is impossible to estimate in a real-world).

\subsection{Analysis}
\subsubsection{Validation of SynthCal}
We conducted an extensive assessment of three distinct camera calibration algorithms, taking into account both rectilinear and wide-angle camera configurations. The dataset employed in this study was generated using a 9$\times$12 checkerboard pattern, and the findings are summarized in \cref{table2}. Our analysis indicates that Bouget's method outperforms both Zhang's and Tsai's methods. This outcome is consistent with previous research, specifically validating the established efficacy of various calibration methods reported by Zollner et al. \cite{zollner2004comparison}.
Upon examining the table, it becomes apparent that calibration accuracy decreases in the case of wide-angle lenses depicted in \cref{fig:4}. This observation aligns with expectations, considering the inherent complexities associated with wide-angle lenses compared to rectilinear lenses. This discovery further underscores the reliability of our synthetic benchmark, demonstrating that center-based patterns are more effective than edge-based patterns in challenging environments. However, the increased complexity of wide-angle lenses negatively impacts their performance compared to edge-based patterns.
The Charuco pattern, achieving the highest score, demonstrates its robustness to noise compared to other patterns, indicating its alignment with real-world data."

\subsubsection{Monocular Configuration}
In the context of monocular settings, our objective was to assess the effectiveness of various camera calibration patterns with different calibration algorithms. We conducted a comprehensive analysis by calculating both the $\mathrm{RPE_{RMS}}$ and $\mathrm{RMSE}$ for all eight available configurations using Zhang's method, as detailed in \cref{table3}.
Our observations revealed that under normal conditions, the Charuco pattern consistently yielded the best results across various camera calibration algorithms. Interestingly, Bouguet's method consistently exhibited the least amount of error, regardless of the pattern type. The combination of both methods produced optimal results in terms of $\mathrm{RPE_{RMS}}$ and $\mathrm{RMSE}$ metrics.
In the presence of noise, our findings indicated that irrespective of the calibration pattern type, Zhang's method outperformed Bouguet's method. This distinction was particularly evident when considering $\mathrm{RMSE}$ metrics, as opposed to the traditional $\mathrm{RPE_{RMS}}$ metrics. We attribute this phenomenon to algorithmic differences and the robustness of calibration algorithms to noise.
These insights highlight the importance of considering $\mathrm{RMSE}$ metrics when evaluating the performance of camera calibration patterns and algorithms in monocular settings.

\begin{figure}[!t]
\begin{center}
\includegraphics[width=0.8\linewidth]{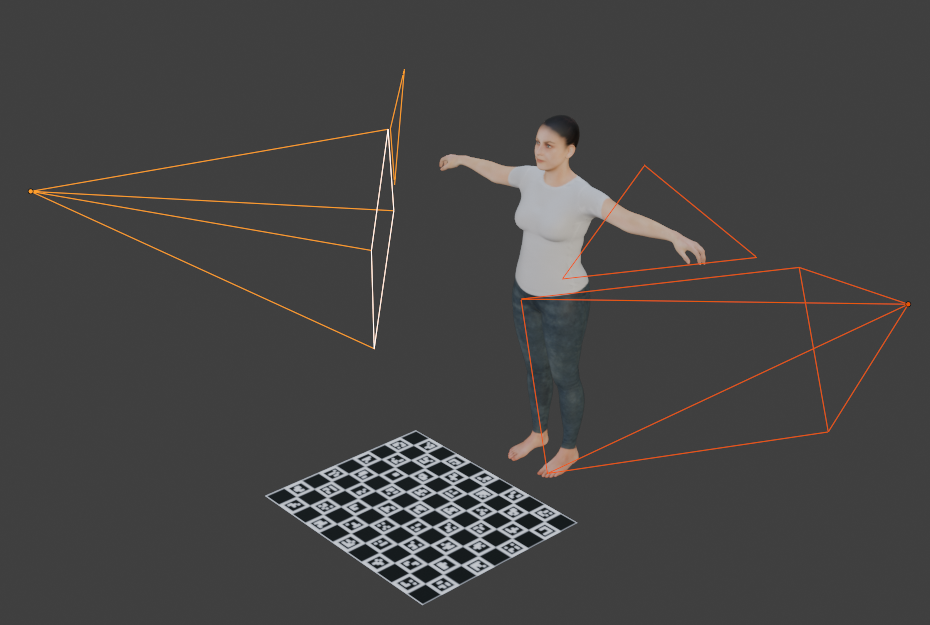}
\end{center}
   \caption{Using SynthCal pipeline with DMCB \cite{ray2023selecting} and EasyMocap \cite{dong2021fast} to estimate 3D pose from multiple view points
   }
\label{fig:5}
\end{figure}

\begin{table*}[!t]
  \begin{center}
  \caption{ MPJPE calculated by comparing the original 24 joint SMPL pose with the predicted SMPL pose using EasyMocap \cite{dong2021fast} for different calibration configurations.}
  \label{table5}
    {\small{
\begin{tabular}{c|c|c|c|c}
\hline
&& Zhang's Method & Tsai's Method & Bouguet's Method \\
\hline
Cameras & Environment &  \multicolumn{3}{c}{MPJPE} \\
\hline
2$\times$Rectilinear & Clean & 5.974 & 6.011 & \textbf{5.967}\\
lens. & Noisy & \textbf{6.313} & 6.381 & 6.344\\
\hline
2$\times$Wide angle & Clean & 5.986 & 6.028 & \textbf{5.962}\\
lens. & Noisy & \textbf{6.440} & 6.485 & 6.478\\
\hline
Rectilinear \& wide& Clean & 6.212 & 6.343 & \textbf{6.413}\\
angle lens & Noisy & \textbf{6.511} & 6.595 & 6.620\\
\hline
\end{tabular}
}}
\end{center}
\end{table*}

\subsubsection{Multi-camera Configuration}
Due to the availability of the absolute position of the calibration pattern, we employed \(RMSE_{cal}\) to quantitatively assess the accuracy of various camera setups and calibration algorithms in multicamera settings, as outlined in \cref{table4}.
Across all camera setups in clean environments, Bouget's method consistently outperformed both Zhang's and Tsai's methods. However, in noisy setups, Zhang's method exhibited greater accuracy compared to Bouget's method. Interestingly, in some camera setups involving wide-angle lenses, Bouget's method performed worse than Tsai's method, reaffirming our earlier observation of Bouget's method's lack of robustness in noisy conditions.
Additionally, our observations revealed that, for stereo setups, identical camera pairs demonstrated superior performance compared to non-identical pairs. This trend persisted even when wide-angle lenses, known for their complexity, were involved. Surprisingly, the combination of two rectilinear lenses consistently outperformed setups comprising one wide-angle and one rectilinear lens.

To further validate our model and assess the accuracy of 3D pose estimation across various camera configurations and calibration algorithms, we utilized the DMCB \cite{ray2023selecting} to simulate a textured human mesh in SMPL \cite{loper2023smpl} format using the motion imported from TotalCapture dataset \cite{trumble2017total} and texture imported from SMPLitex \cite{casas2023smplitex} within Blender as visualized in \cref{fig:5}. This simulated mesh was then captured by multiple cameras positioned at different angles, employing different calibration algorithms and calibration patterns using SynthCal. The rendered videos are then given as input to EasyMocap \cite{dong2021fast} to estimate the 24-joint SMPL pose.

Unlike real-world scenarios where ground truth pose data might be unavailable, here we have access to the real ground truth of the 3D pose, as it was used to create the animated mesh. By comparing the estimated 3D poses with this ground truth, measured through metrics like Mean Per Joint Position Error (MPJPE) which is defined as:

\[
MPJPE = \frac{1}{N}\sum_{i=1}^{N} \sqrt{\sum_{j=1}^{J} \| \mathbf{P}_{ij} - \mathbf{P}_{ij}^{GT} \|_2^2}
\]

where \(N\) is the number of frames, \(J\) is the number of joints, \(\mathbf{P}_{ij}\) denotes the estimated joint position, and \(\mathbf{P}_{ij}^{GT}\) represents the ground truth joint position, we were able to validate our findings.

The results given in \cref{table5} follow a similar trend already established in \cref{table4} hence validating our benchmark. Our results underscored the impact of calibration patterns and algorithms on the accuracy of 3D pose estimation models, although not to a very high extent. This validation highlights the importance of meticulous calibration procedures and algorithm selection in enhancing the accuracy and reliability of machine learning models for tasks like 3D pose estimation.

\section{Conclusion}
\label{sec:conclusion}
In this paper, we introduced the SynthCal pipeline evaluating camera calibration methods. Our research underscores the efficacy of the Charuco pattern coupled with Bouguet's method under standard conditions, while Zhang's method demonstrates superiority in noisy environments. Bouguet's approach fares admirably in pristine setups but encounters challenges with wide-angle lenses and noise. Consistency among camera pairs surpasses mixed configurations, underscoring the significance of uniformity. Our results demonstrated the importance of considering diverse metrics in calibration assessment. 

For future work, we could expand SynthCal to incorporate non-planar calibration algorithms, thereby enhancing its relevance across various camera models and applications.

\subsubsection{Acknowledgements} \label{sec:ack}
The research reported in this paper was supported by the BMBF in the project VidGenSense (01IW21003). 

\bibliographystyle{splncs04}
\bibliography{refs}

\end{document}